\documentclass[sn-mathphys-ay]{sn-jnl}% Math and Physical Sciences Author Year Reference Style
\usepackage{hyperref}
\usepackage{graphicx}%
\usepackage{multirow}
\usepackage{amsmath,amssymb,amsfonts}%
\usepackage{mathrsfs}%
\usepackage[title]{appendix}%
\usepackage{xcolor}%
\usepackage{textcomp}%
\usepackage{manyfoot}%
\usepackage{booktabs}%
\usepackage{algorithm}%
\usepackage{algorithmicx}%
\usepackage{algpseudocode}%
\usepackage{listings}%
\usepackage{float}
\usepackage{lmodern}
\usepackage{anyfontsize}
\usepackage{fix-cm}
\usepackage{xspace}
\usepackage{tcolorbox}
\usepackage{adjustbox}
\usepackage{tabulary}
\usepackage{times}
\usepackage{latexsym}
\usepackage{microtype}

\usepackage{inconsolata}
\usepackage{longtable}
\usepackage{pdfpages}
\begin{document}

\title{AILQA: Evaluating AI-Driven Legal Question Answering Systems for the Indian Legal System}

%%=============================================================%%
% GivenName	-> \fnm{Joergen W.}
% Particle	-> \spfx{van der} -> surname prefix
% FamilyName	-> \sur{Ploeg}
% Suffix	-> \sfx{IV}
% \author*[1,2]{\fnm{Joergen W.} \spfx{van der} \sur{Ploeg} 
%  \sfx{IV}}\email{iauthor@gmail.com}
%%=============================================================%%

\author*[1,4]{\fnm{Shubham Kumar} \sur{Nigam}}\email{s.k.nigam@bham.ac.uk}
\equalcont{These authors contributed equally to this work.}

\author[1]{\fnm{Shubham Kumar} \sur{Mishra}}\email{skmishra20@cse.iitk.ac.in}
\equalcont{These authors contributed equally to this work.}

\author[2]{\fnm{Noel} \sur{Shallum}}\email{noelshallum@gmail.com}

\author[3]{\fnm{Kripabandhu} \sur{Ghosh}}\email{kripaghosh@iserkol.ac.in}

\author[1]{\fnm{Arnab} \sur{Bhattacharya}}\email{arnabb@cse.iitk.ac.in}

\affil*[1]{\orgdiv{Computer Science and Engineering}, \orgname{Indian Institute of Technology}, \orgaddress{\city{Kanpur}, \postcode{208016}, \state{Uttar Pradesh}, \country{India}}}

\affil[2]{\orgdiv{Law}, \orgname{Symbiosis Law School}, \orgaddress{\city{Pune}, \postcode{411014}, \state{Maharashtra}, \country{India}}}

\affil[3]{\orgdiv{Computational and Data Sciences Department}, \orgname{Indian Institute of Science Education and Research}, \orgaddress{\city{Kolkata}, \postcode{741246}, \state{West Bengal}, \country{India}}}

\affil[4]{\orgdiv{School of Computer Science}, \orgname{University of Birmingham}, \orgaddress{\state{Dubai}, \country{UAE}}}

%%%%%%%%%%%%%%%%%%%%%%%%%%%%%%%%%%%%%%%%%%%%

\abstract{This comprehensive study introduces an advanced \textbf{A}rtificial Intelligence for \textbf{I}ndian \textbf{L}egal \textbf{Q}uestion \textbf{A}nswering or \textbf{\texttt{AILQA}} system tailored for the Indian legal context. \texttt{AILQA} leverages a variety of embedding and generative models, including the latest Large Language Models (LLMs), to address the unique challenges posed by the intricate and diverse nature of Indian legal texts, to enhance the accuracy and reliability of legal question responses. We conducted rigorous evaluations using both lexical and semantic metrics that are enriched by expert legal feedback to ensure relevance and accuracy.
Our findings underscore the effectiveness of the Retrieval-Augmented Generation (RAG) paradigm in improving answer quality, particularly in complex legal domains. Additionally, we explored the performance on standardized tests such as the All India Bar Exam (AIBE), thus providing a robust benchmark for a practical application. Under the study’s evaluation protocol, some AI-generated responses received higher ratings than the available reference answers, particularly when they contained accurate and relevant supporting detail. This finding is specific to the evaluated dataset and rating criteria and should not be interpreted as evidence that the models generally outperform qualified legal professionals.
We also discuss the challenges encountered, such as the need for precise context and the risks of model hallucination, and propose directions for future research to further refine AI capabilities in the legal field. This study aims to pave the way for enhanced legal decision-making support systems, making them more accessible and effective for legal professionals and the public alike.}

\keywords{Legal Question Answering, Retrieval Augmented Generation (RAG), Large Language Model (LLM), Embedding Model, QA Model, Indian Legal Domain, Legal Expert Rating}

\maketitle

\section{Introduction}
Question Answering (QA) is an artificial intelligence (AI) task that utilizes Natural Language Processing (NLP) to interpret and respond to queries in natural language, mimicking human interaction \citep{allam2012question, choi2018quac}. Recent advancements in deep learning technologies, particularly with models like Generative Pretrained Transformer 3 (GPT-3) and BERT, have significantly enhanced the capabilities of QA systems in extracting relevant information from large, unstructured datasets \citep{devlin2018bert, qu2019bert, wang2019multi, kassner2020bert}. These systems are increasingly utilized across diverse domains such as healthcare, customer service, and education, leading to substantial improvements in information processing efficiency and service delivery.

In the legal domain, the implementation of QA systems has provided innovative solutions, especially in countries with advanced AI infrastructures like the USA, UK, and Brazil. In these jurisdictions, legal QA systems have been integrated into court proceedings and public legal services, aiding in case summarization, evidence assessment, and even predictive judgments. For instance, in the USA, several states are experimenting with AI to assist in minor traffic violation adjudications and initial hearing assessments \cite{king2020artificial}. Similarly, in the UK, AI is being piloted to navigate legal documents and suggest relevant prior cases to lawyers and judges \cite{collenette2023explainable, drake2022legal}, thereby streamlining legal research efforts.

Despite these advancements, the adoption of AI in the Indian legal system remains in a rudimentary stage. Unlike its Western counterparts, where AI-driven systems are gradually becoming integral to legal operations, India has yet to fully embrace the potential of AI technologies in the judiciary and legal education. The complex legal landscape of India, characterized by a diverse amalgamation of civil, criminal, common, and customary laws, presents unique challenges that have slowed AI integration. Furthermore, the use of English as the primary language for higher judiciary and legal documentation adds another layer of complexity, making the development of effective QA systems more challenging.

Our study aims to bridge this gap by exploring the application of QA models specifically designed for the Indian legal domain. By focusing on criminal cases processed in English due to resource constraints, we aim to demonstrate the potential of AI to transform legal QA within the Indian context. This research not only addresses the technical challenges but also highlights how such innovations could benefit the Indian judiciary and law students alike by providing enhanced access to legal information and fostering a more efficient legal process.

\begin{figure*}[t]
    \centering
    \includegraphics[width=\linewidth]{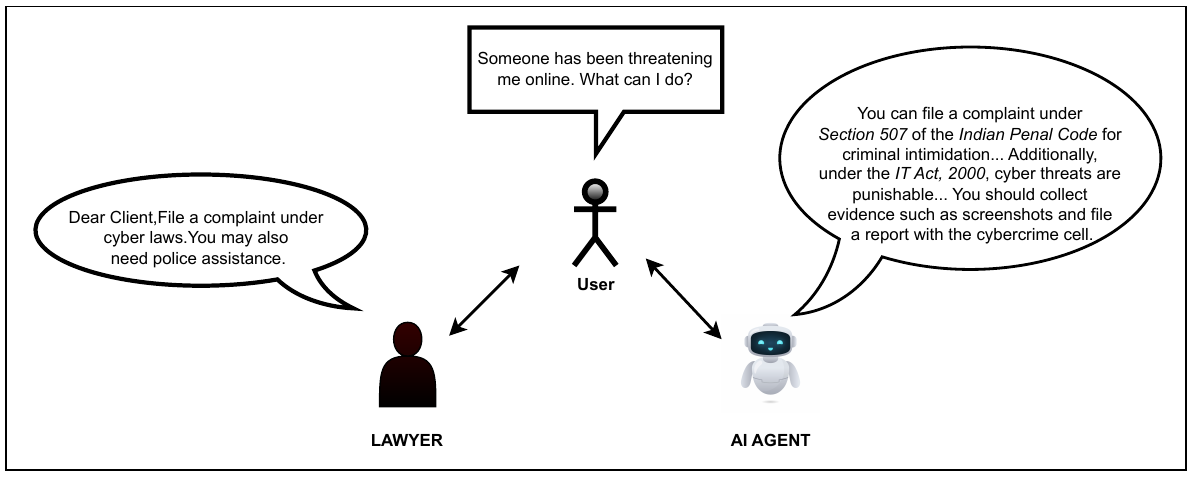}
    \caption{Comparison of a concise reference response and a more detailed AI-generated response to an online-threat query. The example illustrates differences in response detail and structure; greater detail should not, by itself, be interpreted as greater legal correctness.}
    \label{fig:example}
\end{figure*}

Figure \ref{fig:example} illustrates the comparative effectiveness of AI in providing legal advice. A user seeking advice on online threats receives distinctly different responses from a lawyer and an AI agent. The lawyer provides a brief, to-the-point response, advising the user to file a complaint under cyber laws and consider police assistance. Meanwhile, the AI agent offers a more detailed solution, referencing Section 507 of the Indian Penal Code for criminal intimidation and the IT Act, 2000 for cyber threats, along with practical steps like gathering evidence and reporting to the cybercrime cell. While the lawyer's response is concise and less descriptive. However, with the right knowledge base and well-constructed prompts, AI has the potential to offer legal advice of consistent quality, comparable to that given by an experienced lawyer. This suggests that AI can help standardize legal responses, ensuring they are as thorough and accurate as human-provided advice, regardless of the lawyer's approach.

%%%%%%%%%%%%%%%%%%%%%%%%%%%%%%%%%%%%%%%%%%%%%%%%%%%%%%%%%%%%
%%%%%%%%%%%%%%%%%%%%%%%%%%%%%%%%%%%%%%%%%%%%%%%%%%%%%%%%%%%%
The contributions of our study are significant, as they pave the way for the introduction of sophisticated AI tools in an environment where such technological advancements are yet nascent. The following specific model combinations and evaluation strategies employed in our research illustrate the potential of AI to exceed the capabilities of traditional legal analysis tools in some scenarios, thereby setting a precedent for further AI adoption in Indian legal practices:

\begin{itemize}
    \item \textbf{Introduction of RAG in Indian Legal Question-answering:} We pioneer the application of the retrieval-augmented generation (RAG) paradigm for the Indian legal system. This novel contribution involves developing and integrating RAG systems specifically tailored to handle the complexities of India’s diverse legal framework. By enabling the model to dynamically retrieve and utilize relevant past legal cases and statutes as part of the answer generation process, we enhance the model’s ability to produce more accurate, contextually relevant answers.
    \item \textbf{Dataset and Code Availability:} We contribute to the research community by making the AILQA dataset and prediction models publicly available, thereby promoting transparency and reproducibility in legal AI research.
    \item \textbf{Exploration of Embedding Models:} We investigate various combinations of embedding and QA models tailored for legal question answering, and demonstrate their effectiveness in the Indian legal domain.
    \item \textbf{Comprehensive Evaluation Methodology:} We establish a robust evaluation framework that incorporates both lexical and semantic metrics, along with expert legal feedback, to assess the quality of generated answers.
    \item \textbf{Automated Evaluation Framework:} We implement an LLM-based auto-evaluation method to augment human expert judgment and, thus, significantly expedite the evaluation process while maintaining accuracy.
    \item \textbf{Statistical Analysis:} We perform rigorous statistical tests to validate the performance of different models and their outputs, to validate the experimental results statistically.
\end{itemize}

Additionally, we analyze the impact of retrieval-augmented generation (RAG) on answering various types of questions from our legal dataset, by visualizing the results through histograms. By providing concrete examples of hallucinations in generative models such as LLaMA2-70b and GPT-3.5 Turbo, we aim to better understand the limitations of these models in the legal domain. To support further research and ensure reproducibility, we have made the AILQA dataset and the code for our prediction and explanation models publicly available via an GitHub link\footnote{\url{https://github.com/ShubhamKumarNigam/AILQA}}.

\section{Related Work}

In recent years, significant improvements in question-answering (QA) systems have been driven by advances in machine learning models and natural language processing techniques. Introducing models like BERT (Bidirectional Encoder Representations from Transformers) \citep{devlin2018bert} and GPT (Generative Pre-trained Transformer) \citep{floridi2020gpt} has revolutionized the field, enabling systems to understand and process natural language with unprecedented accuracy. Studies such as \citep{devlin2018bert} and \citep{radford2019language} have demonstrated the effectiveness of these models in general QA tasks across various domains, setting a new standard for AI-driven interaction.

Following these foundational models, researchers have explored specific adaptations and enhancements to tailor QA systems for specialized applications. For instance, \citet{yang2018enhancing} introduced models that incorporate external knowledge bases to improve the contextuality of answers in knowledge-intensive tasks.

The legal domain has been a prominent area of research for AI-driven solutions, particularly for tasks such as Legal Judgment Prediction (LJP) and question answering. Legal Judgment Prediction~\citep{strickson2020legal,xu2020multi,feng2023criminal} focuses on predicting decisions of legal cases. Research has explored various machine learning models to perform LJP across jurisdictions such as the European Union, China, and France~\citep{xu2020multi}. In the Indian context, LJP has been attempted by developing specialized datasets like the ILDC corpus, and utilizing hierarchical transformer models for judgment prediction~\citep{malik-etal-2021-ildc,nigam2023nonet,nigam-etal-2024-legal}. This work also emphasizes providing explanations alongside predictions, an aspect crucial for transparency and trust in legal AI systems.

Recent advances have seen the application of transformer models to specific subdomains of LJP, such as using only case facts for prediction~\citep{nigam2023fact}. Additionally, the performance of large language models (LLMs) on legal corpora like ILDC has been studied, showing promising results for improving legal judgment prediction~\citep{vats2023llms}. Similar approaches have been explored in other regions, such as Romania~\citep{masala2021jurbert} and Korea~\citep{hwang2022multi}, where country-specific legal corpora and benchmarks are used to develop more specialized models.

Beyond LJP, question answering based on retrieval-augmented generation (RAG) has gained traction for its ability to effectively combine retrieval mechanisms with generative models to answer complex queries. One significant contribution in this domain is RAG-QA Arena~\citep{han2024rag}, which evaluates domain robustness for long-form RAG-QA. The authors present Long-form Robust QA (LFRQA), a dataset that integrates multiple short extractive answers into coherent, long-form narratives across seven different domains. RAG-QA Arena further leverages model-based evaluators to benchmark the cross-domain generalization of QA systems, offering a robust evaluation platform for future RAG-QA research. This work is especially relevant for evaluating the effectiveness of generative models in complex, multi-document scenarios and has shown that existing LLMs struggle to outperform human-annotated long-form answers.

The dynamic nature of document relevance in RAG-based systems is further explored in the recent work of~\citep{hei2024dr}, which introduces Dynamic-Relevant Retrieval-Augmented Generation (DR-RAG). This two-stage retrieval framework aims to improve document retrieval recall and the accuracy of answers by addressing the limitations of traditional RAG frameworks. DR-RAG proposes a novel approach to mining relevance from both highly relevant (static) documents and lower-relevance (dynamic) documents that may still be crucial to generating accurate answers. By incorporating a classifier to optimize document retrieval and minimize redundant information, DR-RAG achieves substantial improvements in multi-hop question answering accuracy and recall. The study demonstrates the efficacy of DR-RAG across various multi-hop QA datasets, showing enhancements in recall by 86.75\% and improvements in accuracy, exact match (EM), and F1 score by 6.17\%, 7.34\%, and 9.36\%, respectively.

Moreover, the work of~\citep{muludi2024retrieval} introduces the use of RAG combined with GPT-3.5 for processing large external documents. This study focuses on using RAG to improve the accuracy of document-based question answering by leveraging retrieval from external knowledge sources to complement the generative capabilities of language models. The system processes documents automatically and answers user queries by generating responses based on the retrieved content. The study's contributions include the creation of a dataset and performance testing with the Stanford Question Answering Dataset (SQuAD). It demonstrated the superiority of RAG, achieving significant improvements across various metrics such as ROUGE, BERTScore, BLEU, and Jaccard Similarity. This advancement highlights the potential of RAG in real-world applications, especially in mitigating hallucinations and improving the reliability of AI-driven question-answering systems.

Additionally, \citep{wiratunga2024cbr} introduced CBR-RAG, a case-based reasoning approach for retrieval-augmented generation in large language models for legal question answering. This work combines the strengths of case-based reasoning and RAG to effectively leverage past cases and external knowledge sources in generating accurate and relevant answers to legal queries.

While these studies focus on legal judgment prediction, long-form question answering, and improving retrieval strategies, fewer works address the challenge of query-based legal question answering using explainable generative AI models. The task of answering legal questions is closely related to LJP but requires additional capabilities to understand user queries and retrieve relevant legal content.

% In our work, we address these challenges by employing several approaches to evaluate the generated answers in the legal domain thoroughly. Our approach includes both lexical analysis (ROUGE, BLEU) and semantic evaluation using MPNET to determine the contextual similarity between the generated answers and ground truth. We further conduct expert evaluations by having actual legal professionals compare the AI-generated answers with those provided by human lawyers. To streamline the evaluation process, we introduce an auto-evaluation method that leverages large language models (LLMs) to complement human expert evaluation, significantly speeding up the process. We also perform statistical tests, using p-values for pairwise comparisons based on MPNET similarity scores across different experimental conditions. 

% Additionally, we analyze the impact of retrieval-augmented generation (RAG) on answering different types of questions from our legal dataset, visualizing the results through histograms. By providing concrete examples in the hallucination section, we showcase instances where generative models such as LLaMA2-70b, GPT-3.5 Turbo, and others hallucinate, helping us better analyze and understand the shortcomings of these models in the legal domain.

\section{Dataset}
\label{dat:dataset_description}

\subsection{Dataset Compilation}
Our dataset is a robust aggregation of various legal documents essential for training and evaluating the Legal Question Answering system. This collection encompasses statutory texts, judicial decisions, and authoritative legal commentaries specific to the Indian jurisdiction. We sourced statutory laws and regulations directly from the IndiaCode\footnote{\url{https://www.indiacode.nic.in/}}, ensuring the inclusion of all relevant acts listed in Table~\ref{list_of_acts}. Judicial opinions from the Supreme Court from the years 1947 to 2020 were meticulously compiled from IndianKanoon\footnote{\url{https://indiankanoon.org/}}, a comprehensive database offering access to a wide range of legal documents. Furthermore, to enrich the dataset with diverse legal discussions and analyses, we extracted articles and blogs from platforms such as Mondaq\footnote{\url{https://www.mondaq.com/5/India/Criminal-Law}} and LawyersClubIndia\footnote{\url{https://www.lawyersclubindia.com/articles/}}. These sources provide contemporary interpretations and practical insights into criminal law, enhancing the dataset’s relevance for current legal practices.

\subsection{Dataset Statistics and Preprocessing}
The dataset consists of approximately 7,221 legal documents, including both case law and statutory materials. In the preprocessing phase, the dataset underwent rigorous cleaning processes to ensure the integrity and usability of the data. This included the removal of non-essential elements such as headers, footers, extraneous spaces, and line breaks, which could interfere with the text processing algorithms. The cleaned documents were then analyzed to provide statistical insights, as detailed in Table \ref{tab:dataset_stats}. This table presents a breakdown of the document types, showcasing the diversity and volume of the content within the dataset, from judicial rulings to legislative texts and insightful legal articles. These preprocessing steps were crucial in standardizing the dataset for subsequent use in training and testing the AI models, ensuring consistency and reliability in the data fed into our machine learning pipelines.

%%%%%%%%%%%%%%%%%%%%%%%%%%%%%%%%%%%%%%%%%%%%%%%%%%%
\begin{table}[t]
\begin{center}
%\tiny
\resizebox{0.70\columnwidth}{!}
{%
\begin{tabular}{lrr}
\toprule
% \textbf{Data} & \textbf{\thead{Average Word\\Count}} & \textbf{\thead{Number of\\Documents}} \\
\textbf{Data} & \textbf{Word Count (Avg.)} & \textbf{No. of Documents} \\
\midrule
% \makecell{Criminal\\Judgements} & 4021 & 6942 \\ 
Judgements & 4021 & 6942 \\
%\hline
Acts & 28705 & 15 \\
%\hline
Articles & 1557 & 264 \\
\bottomrule
\end{tabular}
}
\caption{\centering Statistical overview of various Criminal Law document distributions}
\end{center}
\vspace*{-2mm}
\label{tab:dataset_stats}
\end{table}
%%%%%%%%%%%%%%%%%%%%%%%%%%%%%%%%%%%%%%%%%%%%%%%%%%%

\subsection{Test Data}
To evaluate the performance of various answer generation and document retrieval models within our legal QA system, we compiled two test datasets from the VidhiKarya website\footnote{\url{https://vidhikarya.com/free-legal-advice}}. Test Set 1 includes 50 legal queries, while Test Set 2 comprises 100 QA pairs. These datasets contain legal queries with expert responses covering five key legal topics: Anticipatory Bail, Criminal Law, Cyber Crime, Juvenile Issues, and Sex Crimes. In Test Set 2, each category contains 20 QA pairs, randomly selected to represent the domain comprehensively. The answers provided by legal experts on VidhiKarya serve as our ground truth, enabling a direct comparison between the generated answers and expert responses. Each question is paired with its corresponding expert answer, facilitating a straightforward evaluation of our models' performance.

%%%%%%%%%%%%%%%%%%%%%%%%%%%%%%%%%%%%%%%%%%%%%%%%%%%
\begin{table}[ht]
\centering
% \tiny
\resizebox{0.7\columnwidth}{!}{%
\begin{tabular}{rl}
\toprule
\textbf{S. No.} & \textbf{Act} \\
\midrule
1 & Indian Penal Code \\
2 & Protection of Children from Sexual Offences Act \\
3 & Criminal Procedural Code \\
4 & Indian Evidence Act \\
5 & Arms Act \\
6 & Information Technology Act \\
7 & Narcotic Drugs and Psychotropic Substances Act \\
8 & Contempt of Courts Act \\
9 & Unlawful Activities Prevention Act \\
10 & Prevention of Money Laundering Act \\
11 & Criminal Procedure Identification Act \\
12 & Extradition Act of 1962 \\
13 & Prisons Act of 1894 \\
14 & Prevention of Corruption Act of 1988 \\
15 & Gram Nyayalayas Act of 2008 \\
\bottomrule
\end{tabular}
}
\caption{\centering List of Acts used as contextual data in the question-answering system.}
\label{list_of_acts}
\end{table}

\subsubsection{All India Bar Exam (AIBE) Dataset}
The All India Bar Exam (AIBE) dataset, published by \cite{tiwari2024aalap}, serves as a critical component to verify our methodology. This dataset encompasses questions and answers from AIBE exams conducted over the past 12 years, from AIBE 4 to AIBE 16. It includes 1,158 multiple-choice questions covering various areas of law, which were manually verified for correctness. The questions primarily test recall abilities, with a few assessing legal reasoning skills. The minimum passing percentage for this exam is set at 40\%, providing a robust benchmark for evaluating the effectiveness of our AI models in a standardized testing environment.

\section{Methodology}
\label{sec:experimental_setup}
\subsection{System Overview}
\begin{figure*}[t]
    \centering
    \includegraphics[width=\linewidth]{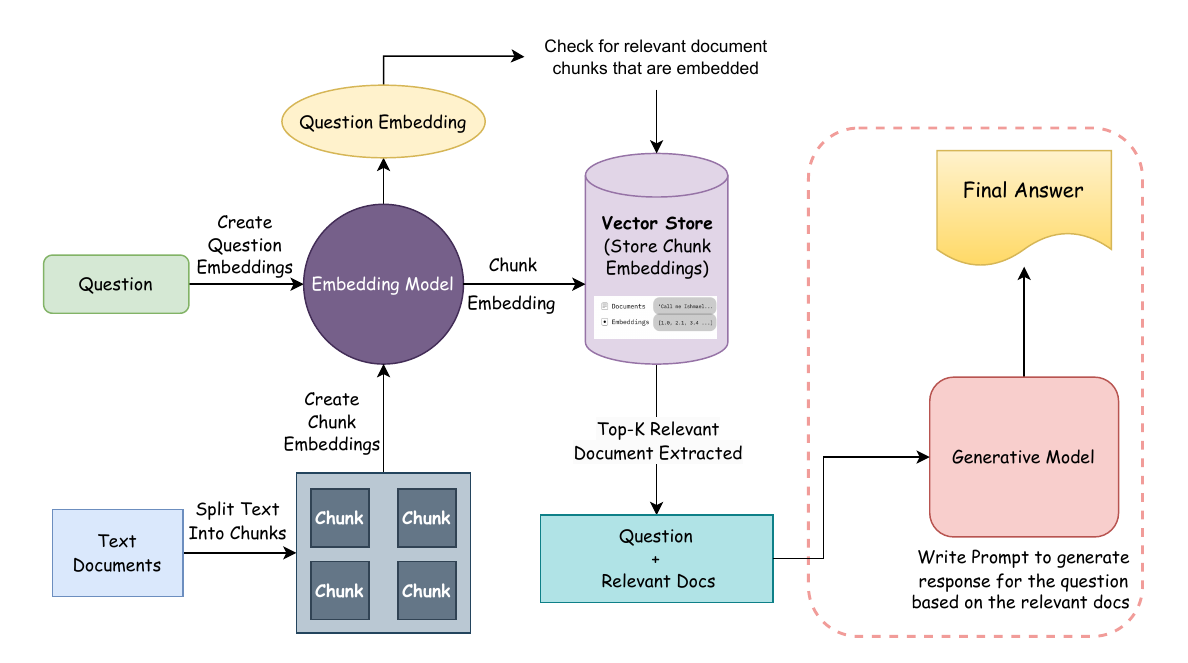}
    \caption{Flowchart illustrating the Legal QA System, detailing the use of GPT-3 Ada and Instructor XL for context extraction, and GPT-3 (Davinci), Flan-UL2, and Llama2-70B for response generation.}
    % \vspace*{-2mm}
    \label{fig:Flowchart}
\end{figure*}

Our Legal Question Answering (QA) system employs an advanced Retrieval-Augmented Generation (RAG) architecture that significantly enhances the capabilities of generative AI models. This methodology integrates dynamic retrieval of contextual information from a comprehensive legal database, which substantially augments the generative process. The RAG system is specifically tailored to address the complex needs of legal QA by providing precise, contextually relevant answers to legal queries and facilitating rigorous examination preparations for the legal bar exam. Figure \ref{fig:Flowchart} provides a visual representation of our QA system's workflow. This figure outlines how documents are stored in a database following various preprocessing steps and how relevant document chunks are retrieved to answer user queries or legal bar exam questions. When a question is posed, it is combined with the retrieved chunk and passed to a generative model, accompanied by a defined prompt, to produce the answer.

%%%%%%%%%%%%%%%%%%%%%%%%%%%%%%%%%%%%%%%%%%%%%%%%%%%%%%%%%%%%
%%%%%%%%%%%%%%%%%%%%%%%%%%%%%%%%%%%%%%%%%%%%%%%%%%%%%%%%%%%%

We have discussed the processes involved in making this architecture work for our tasks in the sections below.

\subsection{Chunking of Documents}
\label{subsec:chunking_documents}

To handle large legal documents efficiently within the computational limits of language models, we implemented a chunking strategy. This approach divides extensive texts into manageable pieces, each consisting of 2000 characters, with an overlap of 250 characters to maintain narrative continuity. This method ensures that the input to our language models remains within their token size limits and preserves the context necessary for generating accurate responses. We utilized LangChain's \texttt{CharacterTextSplitter}\footnote{\url{https://python.langchain.com/v0.1/docs/modules/data_connection/document_transformers/character_text_splitter/}} for this process, optimizing the chunk size to balance between model capacity and contextual completeness.

\subsection{Embeddings Creation and Vector Store Database}

In the previous section, we discussed how we divided large documents in the dataset into smaller chunks of textual data. Now, we use embedding models to create embeddings for each chunk and store these embeddings in the form of a vector database. This approach is helpful in retrieving the most contextually relevant chunks quickly and efficiently.

We have used \texttt{ChromaDB}\footnote{\url{https://python.langchain.com/v0.2/docs/integrations/vectorstores/chroma/}}, an open-source database developed by Chroma, for storing and using vector embeddings. The Langchain framework supports an easily integratable pipeline for creating the vector database with the help of ChromaDB. The chunks obtained in the previous section are passed along with the embedding model in the pipeline, and the path for saving the embedding database is provided, allowing for seamless database creation at the desired location.

For creating the embeddings, we utilized three models: \texttt{Ada}\footnote{\url{https://platform.openai.com/docs/guides/embeddings}} by OpenAI, \texttt{Instructor-XL} \citep{su2022one}, and \texttt{Mxbai}\footnote{\url{https://www.mixedbread.ai/blog/mxbai-embed-large-v1}}. We chose these models as they ranked among the top embedding models on the MTEB leaderboard\footnote{\url{https://huggingface.co/spaces/mteb/leaderboard}} available on HuggingFace. Each model differed in size and embedding dimension, leading to the creation of three distinct vector store databases. This allowed us to test the quality of retrieval in our tasks.
Here are some details about the embedding models we used:

\subsubsection{OpenAI's Ada}
The Ada model from OpenAI generates 1536-dimensional embeddings at a cost of \$0.0004 per 1000 tokens. For our dataset of 61.6 million tokens, the total cost for embedding generation was approximately \$24.7. It can be accessed through the API token provided by OpenAI. The documentation for usage can be found here\footnote{\url{https://platform.openai.com/docs/api-reference/embeddings/create}}.

\subsubsection{Instructor-XL}
The Instructor-XL model produces 768-dimensional embeddings, optimized for instruction-based embedding creation and retrieval tasks. It can generate text embeddings tailored to any task (e.g., classification, retrieval, clustering, text evaluation, etc.) or domain (e.g., science, finance, etc.) by simply providing the task instruction in natural language. It is available on the HuggingFace platform\footnote{\url{https://huggingface.co/hkunlp/instructor-xl}}.

\subsubsection{Mxbai}
Mxbai is an English embedding model with an embedding dimension of 1024. It is also available on the Hugging Face platform\footnote{\url{https://huggingface.co/mixedbread-ai/mxbai-embed-large-v1}}.
After creating the embeddings, the chunks are stored with it's metadata and unique id in the vector database.

\subsection{Query Processing and Document Retrieval}

As discussed in the previous section, the chunked data is stored in a vector database. To retrieve relevant chunks for our question-answering tasks, we utilize LangChain's vector data similarity search\footnote{\url{https://python.langchain.com/v0.1/docs/modules/data_connection/vectorstores/}} pipeline. When a query is submitted, the question is converted into embeddings using the same model employed during database creation. The similarity search pipeline then performs a cosine similarity search to identify the most relevant document chunks. In our setup, the top three chunks, based on cosine similarity scores, are used as the context for answering the question.

The implemented RAG architecture should be interpreted as a controlled dense-retrieval baseline rather than a fully optimised contemporary legal RAG pipeline. It relies on fixed-size character chunks, embedding-based cosine similarity, and top-k retrieval without hybrid lexical–semantic retrieval, metadata filtering, query reformulation, or a dedicated reranking stage. This deliberately simple configuration enables a controlled comparison of embedding and generative models, while also allowing us to examine the conditions under which retrieved context improves or degrades legal answer generation.

\subsection{Answer Generation}

Once the relevant context has been retrieved, as described in the previous section, we pass the question along with the context to the generative model to produce a pertinent response. To ensure the quality of the generated answers, we have crafted specific prompts that guide the generative model in generating high-quality responses. We experimented with different models, including Davinci (text-davinci-003)\footnote{\url{https://platform.openai.com/docs/deprecations}}, Llama2-70B \citep{touvron2023llama}, Flan-UL2 \citep{wei2021finetuned}, GPT-3.5 Turbo\footnote{\url{https://platform.openai.com/docs/models/gpt-3-5-turbo}}, Llama3-70B \citep{llama3modelcard} and Mixtral-8x7B \citep{jiang2024mixtral}, and fine-tuned prompts to compare the quality of the responses.

Table \ref{tab:language_models} provides details on maximum token length, pricing, and the prompts used for each model. The table shows that some prompts are tailored to specific models while others are consistent across models. We iteratively refined the prompts by manually evaluating responses to a random set of 10 questions. This iterative process allowed us to ensure the quality of the prompts before applying them to our test data.

\begin{table}[ht]
\centering
\resizebox{\textwidth}{!}{%
\begin{tabular}{ p{0.12\textwidth} p{0.12\textwidth} p{0.10\textwidth} p{0.62\textwidth} }
\toprule
\textbf{Model Name} & \textbf{Max Tokens} & \textbf{Pricing} & \textbf{Prompt} \\
\midrule 
Davinci & 4096 & \$0.02 / 1K tokens & ``Your task is to answer a question as a legal assistant to the best of your abilities, using the context given in the document. If the country is not mentioned in the question, your response should be related to India. You have knowledge of all laws and legal judgments of India. Be detailed in your answer, provide relevant sections and case laws in your answer only if you are confident that they are correct. Note that if you do not know the answer, it is acceptable to say Sorry, I don’t know. Context:\{\} Question:\{\}.'' \\ \midrule
Llama2-70B & 4096 & \$0.65 / 1M tokens & ``You are an honest legal advisor. Your task is to answer a question as a legal assistant to the best of your abilities based on the context provided. If the country is not mentioned in the question, your response should be related to India. You have knowledge of all laws and legal judgments of India. Be detailed in your answer, provide relevant sections and case laws in your response only if you are confident that they are correct. If you are unsure about an answer, truthfully say “I don’t know”. Context:\{\} Question:\{\}'' \\ \midrule
Flan-UL2 & 2048 & N/A & ``Answer the following question using the context by reasoning step by step. If you don’t know the answer, just say Sorry, I don’t know. Context:\{\} Question:\{\}'' \\ \midrule
GPT-3.5 Turbo & 4096 & \$0.0005 / 1K tokens & ``Your task is to answer a question as a legal assistant to the best of your abilities, using the context given in the document. If the country is not mentioned in the question, your response should be related to India. You have knowledge of all laws and legal judgments of India. Be detailed in your answer, provide relevant sections and case laws in your answer only if you are confident that they are correct. Note that if you do not know the answer, it is acceptable to say ``Sorry, I don’t know." Question: \{\} Context: \{\}.'' \\ \midrule
Llama3-70B & 8000 & \$0.65 / 1M tokens & ``Your task is to answer a question as a legal assistant to the best of your abilities, using the context given in the document. If the country is not mentioned in the question, your response should be related to India. You have knowledge of all laws and legal judgments of India. Be detailed in your answer, provide relevant sections and case laws in your answer only if you are confident that they are correct. Note that if you do not know the answer, it is acceptable to say ``Sorry, I don’t know." Question: \{\} Context: \{\}.'' \\ \midrule
Mixtral-8x7B & 32K & \$0.50 / 1M tokens & ``Your task is to answer a question as a legal assistant to the best of your abilities, using the context given in the document. If the country is not mentioned in the question, your response should be related to India. You have knowledge of all laws and legal judgments of India. Be detailed in your answer, provide relevant sections and case laws in your answer only if you are confident that they are correct. Note that if you do not know the answer, it is acceptable to say Sorry, I don’t know." Question: \{\} Context: \{\}.'' \\ 
\bottomrule
\end{tabular}
}
\caption{\centering Specifications of Answer Generation Models: Max Tokens, Pricing, and Prompt Details}
\label{tab:language_models}
\end{table}

\section{Evaluation Metrics}
\label{sec:performance_metrics}

To ensure a comprehensive assessment of our question-answering system, we employed a range of evaluation metrics. These metrics are designed to measure the accuracy, relevance, and reliability of the answers generated by our system:

\begin{enumerate}
    \item \textbf{Lexical Similarity-Based Evaluation:}
    We utilized ROUGE scores \cite{lin-2004-rouge} (ROUGE-1, ROUGE-2, and ROUGE-L) and the BLEU Score \cite{papineni-etal-2002-bleu} to measure the lexical similarity between the generated answers and the reference answers. These metrics evaluate the overlap of n-grams and the sequence of words, providing a quantitative measure of the linguistic quality of the generated text.

    \item \textbf{Semantic Similarity Based Method:}
    To assess the semantic accuracy of the responses, we employed the MPNET base v2 \cite{10.5555/3495724.3497138} sentence transformer model from HuggingFace\footnote{\href{https://huggingface.co/sentence-transformers/all-mpnet-base-v2}{huggingface/sentence-transformers/all-mpnet-base-v2}}. This model projects sentences into a 768-dimensional vector space, enabling us to perform detailed comparisons of semantic closeness between the generated answers and the ground truth.

    \item \textbf{Expert Evaluation:}
    Human evaluation was conducted by three legal evaluators who were third- and fourth-year undergraduate law students enrolled at National Law Universities in India. The evaluators possessed academic training in Indian law and were familiar with legal research, statutory interpretation, and case-law analysis.

    Each generated response was independently evaluated by all three evaluators using a five-point Likert scale based primarily on legal accuracy, relevance, completeness, and the appropriate use of statutory provisions or case law. The evaluators were provided with the legal question, the corresponding reference answer, the generated response, and information regarding the model identity and whether Retrieval-Augmented Generation was used. Therefore, the evaluation was not conducted under blinded conditions.

    After the independent evaluation, responses for which the evaluators assigned differing ratings were discussed to establish a consensus score. Where disagreement remained, the assessment was reviewed with a senior legal expert who is a practising legal professional, and the final rating was determined through discussion and consensus. The resulting consensus ratings were used in the reported expert-evaluation results.

The rating criteria were as follows:

    \textbf{[1]:} The answer is entirely incorrect or fails to provide any answer. 
    
    \textbf{[2]:} The model misunderstood the question and did not offer a relevant response. 
    
    \textbf{[3]:} The answer is partly accurate but overlooks essential details. 
    
    \textbf{[4]:} A comparable, relevant answer to the ground truth. 

    \textbf{[5]:} The answer is legally accurate, directly relevant, sufficiently complete, and supported by appropriate statutory provisions or precedents where applicable. Additional detail is rewarded only when it is correct, relevant, and useful; verbosity alone does not increase the rating.
    
    % \textbf{[5]:} The answer is entirely accurate and relevant, providing a superior response to the expert's answer. 

    \item \textbf{Statistical Comparison of Model Configurations:}
    We conducted pairwise statistical comparisons between the different experimental configurations using the per-question semantic-similarity scores generated by the MPNET model. The resulting p-values were used to assess whether the differences observed between model configurations were statistically significant. A p-value of 0.05 or lower was treated as evidence that the corresponding performance difference was unlikely to have occurred by chance.

    These statistical tests compare the semantic-performance distributions of the model configurations. They should not be interpreted as measures of agreement or consistency among the human legal evaluators.

    \item \textbf{Auto-Evaluation Framework:}
    For an objective assessment of our system, we employed an LLM-based auto-evaluation framework using LangChain's \texttt{load\_evaluator}\footnote{\href{https://python.langchain.com/v0.1/docs/guides/productionization/evaluation/string/scoring_eval_chain/}{LangChain Auto-Evaluation Documentation}}. This framework compares the generated answers to the ground truth using GPT-4 as the evaluation model. It quantifies the relevance of the responses, where a score of 1 indicates high relevance and 0 indicates high irrelevance.
\end{enumerate}

These diverse metrics provide a robust framework for evaluating our system, ensuring that it meets the rigorous standards required for effective application in the legal domain.

\section{Results and Analysis}
This section presents the outcomes of our experiments, which were organized into two distinct phases to evaluate different combinations of generative and embedding models. These phases helped us explore how different models interact and the resultant effects on the quality of the generated answers, particularly focusing on the role of embedding models in enhancing context extraction and the efficacy of generative models in producing accurate responses.

In Phase 1, we utilized Test Set 1, which comprised 50 legal questions, to conduct an initial assessment of our model combinations. This set provided a preliminary understanding of how each model performs under controlled conditions without extensive contextual diversity.

Phase 2 expanded the evaluation to Test Set 2, consisting of 100 legal questions distributed across five categories, as detailed in Section \ref{dat:dataset_description}. This phase was designed to test the robustness of the models in handling a broader array of legal questions and to assess the scalability of the RAG architecture. Additionally, within Phase 2, we also assessed the performance of these models on the Legal Bar Exam Dataset. This part of the evaluation was crucial for determining the effectiveness of the RAG system in a highly specialized legal context.

The following sections provide detailed insights into the results, as well as the various methods employed to understand the effectiveness of the RAG system in generating answers.

\subsection{Results for Legal QA}
\label{sec:results_and_analysis}

Table \ref{tab:average_scores} presents the performance evaluation of various generative models for legal question answering, using a multifaceted approach. The column ``Embedding Model'' specifies the embedding model employed to extract the context for each question. An empty column indicates that the generative model produced the answer without any contextual help.
Below, we discuss the different metric scores presented in the table.

\subsubsection{Lexical Based Evaluation}
We employed ROUGE and BLEU scores to assess the lexical similarity between the generated answers and the reference texts. These metrics provided insights into the precision of word and phrase usage within the generated responses. In Table \ref{tab:average_scores}, for the Test Set 1 dataset, Davinci achieves high scores with and without context for ROUGE-1, ROUGE-2, ROUGE-L, and BLEU scores. The ROUGE and BLEU scores collectively indicate that the Ada model enhances performance when used with Davinci compared to Instructor-XL.

Similarly, for the Test Set 2 dataset, the GPT-3.5 Turbo model achieves better ROUGE and BLEU scores with context using Mxbai as the embedding model.

However, we cannot fully rely on these scores as they do not account for semantic meaning or syntactic structure beyond surface-level word matching. Therefore, in subsequent sections, we discuss other evaluation metrics that include semantic analysis and human evaluation to provide a more comprehensive assessment.

\subsubsection{Semantic Evaluation}
For the semantic-based evaluation, we have used MPNET's ``all-mpnet-base-v2'' \citep{song2020mpnet}\footnote{\url{https://huggingface.co/sentence-transformers/all-mpnet-base-v2}}, which is a sentence-transformer model available on HuggingFace to get the similarity scores based on the semantic similarity between generated answer and the ground truth. The model used here is very capable in catching the semantic information of sentences and used in variety of tasks including the sentence similarity task needed for getting similarity score for our task.

For the Test Set 1, \textbf{Llama2-70B} achieved the highest average MPNET score of \textbf{0.611}. However, Llama2-70B did not perform as well when using context extracted by the Ada or Instructor-XL models, with scores of 0.594 and 0.599, respectively. In contrast, \textbf{Davinci}, when used without context, scored 0.561, and when used with context from Ada and Instructor-XL, secured scores of 0.566 and 0.574, respectively. Lastly, the \textbf{Flan-UL2} model, regardless of the embedding model used, did not show scores as high as Llama2-70B or Davinci. In the case of the Test Set 2, \textbf{Llama3-70B} independently performed well, producing the most semantically similar results compared to the ground truth. However, no model among \textbf{Llama3-70B}, \textbf{Mixtral-8x7B}, and \textbf{GPT-3.5 Turbo} showed a performance boost by using the context, and generally, their performance degraded with context inclusion.

Using MPNET similarity scores, we quantitatively measured the semantic similarity between generated answers and ground truth, providing insights into model performance with and without contextual embedding. However, this approach lacks the nuanced understanding and subjective judgement of human evaluation, which can more accurately capture the quality and relevance of generated answers in a legal context. The limitations of relying solely on automated similarity scores highlight the need for human evaluation to comprehensively assess the performance of generative models in complex tasks like legal question answering.

\subsubsection{Expert Evaluation}
\label{subsec:expert_evaluation}
Legal experts reviewed the answers on a Likert scale from 1 to 5, focusing on accuracy and relevance. The feedback indicated that certain models, especially those with advanced generative capabilities like GPT-3 variants, often produced responses that met or exceeded the quality of reference answers, demonstrating the potential of AI in legal expertise augmentation, as detailed in Table \ref{tab:expert_scores} and average in Table \ref{tab:average_scores}.

%%%%%%%%%%%%%%%%%%%%%%%%%%%
\begin{table}[tb]
\centering
\resizebox{0.60\textwidth}{!}{%
\begin{tabular}{l l *{5}{r}}
\toprule
\textbf{Embedding} & \textbf{Generative} & \multicolumn{5}{c}{\textbf{Rating Score}} \\
\cmidrule{3-7}
\textbf{Model} & \textbf{Model} & \textbf{1} & \textbf{2} & \textbf{3} & \textbf{4} & \textbf{5} \\
\midrule
- & Davinci  & 0 & 9 & 13 & 20 & 8 \\
%\hline
- & Llama2-70B & 1 & 11 & 9 & 15 & 13 \\ 
%\hline
Ada & Davinci & {2} & {7} & {6} & {12} & {21} \\
%\hline
Instructor & Davinci & 2 & 7 & 11 & 15 & 15 \\
%\hline
Ada & Llama2-70B & 0 & 3 & 13 & 33 & 1 \\
%\hline
Instructor & Llama2-70B & 10 & 8 & 7 & 9 & 16 \\
%\hline
Ada & Flan-UL2 & 11 & 33 & 5 & 1 & 0 \\ 
%\hline
Instructor & Flan-UL2 & 5 & 36 & 9 & 0 & 0 \\ 
\bottomrule
\end{tabular}
}
\caption{Ratings from Legal Experts for Different Combinations of Embedding and Generative Models. A `-' in the `Embedding Model' column indicates that no embedding model was used to retrieve context for the corresponding generative model.`}
\label{tab:expert_scores}
\end{table}

%%%%%%%%%%%%%%%%%%%%%%%%%%%%%%%%

\begin{table}[ht]
    \centering
    \resizebox{\columnwidth}{!}{%
    \begin{tabular}{ l l c c c c c c }
        \hline
        \textbf{Embedding} & \textbf{Generative} & \multicolumn{4}{c}{\textbf{Lexical Based Evaluation}} & \textbf{Semantic Evaluation} & \textbf{Expert Evaluation}\\ 
        \cmidrule{3-6}
        \textbf{Model} & \textbf{Model} & \textbf{Rouge-1} & \textbf{Rouge-2} & \textbf{Rouge-L} & \textbf{BLEU} & \textbf{MPNET Score} & \textbf{Rating Score} \\ \midrule
        \multicolumn{8}{ c }{\textbf{Results from Test 1 Dataset}} \\ \midrule
        - & Davinci & \textbf{0.267} & 0.052 & \textbf{0.158} & 0.010 & 0.561 & 3.54 \\ %\hline
        - & Llama2-70B & 0.149 & 0.035 & 0.090 & 0.007 & \textbf{0.611} & 3.50 \\ %\hline
        Ada &  Davinci & 0.242 & \textbf{0.062} & 0.147 & \textbf{0.022} & 0.566 & \textbf{3.74} \\ %\hline
        Instructor-XL & Davinci & 0.229 & 0.053 & 0.139 & 0.016 & 0.574 & 3.68 \\ %\hline
        Ada & Llama2-70B & 0.163 & 0.040 & 0.099 & 0.011 & 0.594 & 3.64 \\ %\hline
        Instructor-XL & Llama2-70B & 0.160 & 0.037 & 0.094 & 0.008 & 0.599 & 3.26 \\ %\hline
        Ada & Flan-UL2 & 0.122 & 0.021 & 0.081 & 0.010 & 0.301 & 1.92 \\ %\hline
        Instructor & Flan-UL2 & 0.121 & 0.013 & 0.081 & 0.001 & 0.343 & 2.08 \\ \midrule
        \multicolumn{8}{ c }{\textbf{Results from Test 2 Dataset}} \\ \midrule
        - & Llama3-70B & 0.20 & 0.05 & 0.11 & 0.09 & \textbf{0.65} & 4.43 \\ %\hline
        Mxbai & Llama3-70B & 0.22 & 0.06 & 0.12 & 0.1 & 0.63 & 3.37 \\ %\hline
        - & Mixtral-8x7B & 0.19 & 0.05 & 0.11 & 0.09 & 0.62 & \textbf{4.59} \\ %\hline
        Mxbai & Mixtral-8x7B & 0.23 & 0.06 & 0.13 & 0.12 & 0.62 & 4.02 \\ %\hline
        - & GPT-3.5 Turbo & \textbf{0.26} &\textbf{ 0.07} & 0.15 & 0.14 & 0.64 & 3.43 \\ %\hline
        Mxbai & GPT-3.5 Turbo & \textbf{0.26 }& \textbf{0.07} & \textbf{0.16} & \textbf{0.16} & 0.62 & 3.55 \\ \bottomrule
    \end{tabular}}
    \caption{This table compares different embedding and generative model combinations across various evaluation metrics for Test 1 and Test 2 datasets. The highest scores in each metric are highlighted in bold. A `-' in the `Embedding Model' column indicates that no embedding model was used to retrieve context for the corresponding generative model.}
    \label{tab:average_scores}
\end{table}
%%%%%%%%%%%%%%%%%%%%%%%%%%%%%%%%%%%%

\subsubsection{Auto-Evaluation Framework}
Finally, we utilized an LLM-based auto-evaluation framework to objectively assess the relevance of the generated answers. This framework allowed for a scalable and consistent evaluation, reinforcing findings from human expert reviews and semantic assessments. Table \ref{tab:max_questions_score1} compares model correctness across 100 evaluation questions. The Llama3-70B model outperformed all others, both with and without the RAG, in producing answers comparable to those of professional lawyers. However, its performance declined with RAG due to suboptimal content extraction causing hallucinations. Conversely, the `Mixtral-8x7B' model performed better with RAG, suggesting that models with less prior knowledge benefit from additional context. Larger models, already trained on extensive data, tend to hallucinate with added context, reducing performance.\\
The LLM-based auto-evaluation framework proved efficient and scalable, reducing the need for human evaluation. It provided consistent scoring and the results highlighted that smaller models benefit from RAG, while larger models may hallucinate with added context.

%%%%%%%%%%%%%%%%%%%%%%%%%%%%%%%%%%%%%%%%%%%%%%%%%%%%%%%%%%%%

\begin{table}[ht]
    \centering
    \resizebox{0.6\columnwidth}{!}{%
    \begin{tabular}{l c c}
        % \hline
        \toprule
        \textbf{Model} & \textbf{RAG} &\textbf{Questions with Score 1} \\
        % \hline
        \midrule
        LLAMA3-70B & NO & \textbf{79} \\
        GPT-3.5-Turbo & NO & 77 \\
        LLAMA3-70B & YES & 75 \\
        Mixtral-8x7B & YES & 73 \\
        Mixtral-8x7B & NO & 71 \\
        GPT-3.5-Turbo & YES & 70 \\
        % \hline
        \bottomrule
    \end{tabular}
    }
    \caption{Table summarizing the auto-evaluated scores by GPT-4 for various models. The table records instances where models achieved a perfect score of 1, indicating complete alignment with the ground truth. The columns list the number of questions where each model scored 1, while the rows categorize the models with or without RAG system.}
    \label{tab:max_questions_score1}
\end{table}

\subsubsection{Statistical Significance Scores}
\label{sec:Statistical-Significance-Table}
Table \ref{tab:p-value-table} shows comparative analysis P-values for pairwise statistical comparisons between different experimental settings based on MPNET similarity scores for Test Dataset 1. The table is symmetric across the diagonal, hence representing in a lower triangular format. The meaningful data (in this case, p-values) are only present in the lower half of the table, below the main diagonal. The main diagonal and the upper half of the table (above the main diagonal) are filled with placeholder symbols ``-''. This means that the comparison of Model A vs. Model B will have the same p-value as Model B vs. Model A, yielding the same statistical significance regardless of the comparison order.

Similarly, Table \ref{tab:p-value_table_2} presents the p-values for Test Dataset 2, following the same format. The ``+'' symbol in both tables indicates the use of Retrieval-Augmented Generation (RAG) for generative models, utilizing embeddings from the model specified after the ``+'' sign.

\begin{itemize}
    \item \textbf{Highly Similar Models:} Several comparisons in both tables, such as `Ada+Flan-UL2' vs. `Ada+Llama2-70B' in Table \ref{tab:p-value-table}, and `Llama3-70B+Mxbai' vs. `Mixtral-8x7B+Mxbai' in Table \ref{tab:p-value_table_2}, show p-values close to 0.0000, indicating extremely high statistical significance. This reflects substantial differences in model performance across different settings, especially when RAG is used with certain models.
    
    \item \textbf{Marginally Significant Comparisons:} Table \ref{tab:p-value-table} has a few comparisons with p-values slightly above 0.05, such as `Instructor+Llama2-70B' vs. `Ada+Davinci' with a p-value of 0.1333, indicating less pronounced differences between models. Similarly, in Table \ref{tab:p-value_table_2}, the comparison `Mixtral-8x7B+Mxbai' vs. `GPT-3.5 Turbo+Mxbai' has a p-value of 0.6591, suggesting a closer similarity in their performance when using RAG.

    \item \textbf{High P-values:} In both tables, some comparisons yield very high p-values, indicating that the differences between models are not statistically significant. For example, in Table \ref{tab:p-value-table}, `Ada+Llama2-70B' vs. `Davinci' has a p-value of 0.7206. Similarly, in Table \ref{tab:p-value_table_2}, `GPT-3.5 Turbo+Mxbai' vs. `Llama3-70B' shows a p-value of 0.1979. These high p-values suggest that these models have similar MPNET similarity scores, and their performance may be considered equivalent.

    \item \textbf{Diversity in Model Performance:} The wide range of p-values across both tables highlights the diversity in model performance. Some models, especially when enhanced with RAG, show significant differences in capabilities, while others demonstrate closer performance in generating legal answers, depending on the configuration and datasets.

    \item \textbf{Importance of Context:} Context remains critical, particularly in legal domains. In both tables, even marginally significant p-values (like 0.0666 for `Llama3-70B+Mxbai' vs. `Mixtral-8x7B+Mxbai' in Table \ref{tab:p-value_table_2}) can suggest performance variations that might be vital in specific legal applications. This reinforces the importance of using embeddings and RAG for certain legal scenarios.

    \item \textbf{Variability in Legal Answering Capabilities:} Both tables reflect the variability in how these models answer legal questions. Some models show clear performance differences, while others are more closely aligned, depending on whether or not RAG is used, as shown by varying p-values across datasets.
\end{itemize}

%%%%%%%%%%%%%%%%%%%%%%%%%%%%%%%%%%%%%%%%%%%%%%%%%%%%%%%%%
\begin{table}[ht]
\centering
\resizebox{\textwidth}{!}{%
\begin{tabular}{lcccccccc}
% \hline
\toprule
 &
  \multicolumn{1}{l}{\textbf{Davinci}} &
  \multicolumn{1}{l}{\textbf{Llama2-70B}} &
  \multicolumn{1}{l}{\textbf{Ada+Davinci}} &
  \multicolumn{1}{l}{\textbf{\begin{tabular}[c]{@{}l@{}}Instructor+\\ Davinci\end{tabular}}} &
  \multicolumn{1}{l}{\textbf{\begin{tabular}[c]{@{}l@{}}Ada+\\ Llama2-70B\end{tabular}}} &
  \multicolumn{1}{l}{\textbf{\begin{tabular}[c]{@{}l@{}}Instructor+\\ Llama2-70B\end{tabular}}} &
  \multicolumn{1}{l}{\textbf{\begin{tabular}[c]{@{}l@{}}Ada+\\ Flan-UL2\end{tabular}}} \\ 
  \midrule
  % \hline

{Llama2-70B}                                                       & 0.0786 & -      & -      & -      & -      & -      & -       \\ 
{\begin{tabular}[c]{@{}l@{}}Ada + Davinci\end{tabular}}           & 0.2527 & 0.0366 & -      & -      & -      & -      & -       \\ 
{\begin{tabular}[c]{@{}l@{}}Instructor + Davinci\end{tabular}}    & 0.4387 & 0.0948 & 0.4596 & -      & -      & -      & -       \\ 
{\begin{tabular}[c]{@{}l@{}}Ada + Llama2-70B\end{tabular}}        & 0.7206 & 0.1237 & 0.2089 & 0.3715 & -      & -      & -       \\ 
{\begin{tabular}[c]{@{}l@{}}Instructor + Llama2-70B\end{tabular}} & 0.4678 & 0.2900 & 0.1333 & 0.2627 & 0.7035 & -      & -       \\ 
{\begin{tabular}[c]{@{}l@{}}Ada + Flan-UL2\end{tabular}}          & 0.0000 & 0.0000 & 0.0000 & 0.0000 & 0.0000 & 0.0000 & -       \\ 
{\begin{tabular}[c]{@{}l@{}}Instructor + Flan-UL2\end{tabular}}   & 0.0000 & 0.0000 & 0.0000 & 0.0000 & 0.0000 & 0.0000 & 0.2451  \\ 
\bottomrule
% \hline
\end{tabular}%
}
\caption{Comparative analysis of p-values for pairwise statistical comparisons between different generative models with and without Retrieval-Augmented Generation (RAG) on Test Dataset 1. In the table,``+" denotes the use of RAG, where the generative model output incorporates embeddings from the specified model following the ``+" sign. P-values are calculated based on MPNET similarity scores.}
\label{tab:p-value-table}
\end{table}
%%%%%%%%%%%%%%%%%%%%%%%%%%%%%%%%%%%%%%%%%%%%%%%%%%%%%%%%%%%%%%%
%%%%%%%%%%%%%%%%%%%%%%%%%%%%%%%%%%%%%%%%%%%%%%%%%%%%%
\begin{table}[ht]
\centering
\resizebox{\textwidth}{!}{%
\begin{tabular}{lccccc}
\toprule
 &
  \textbf{Llama3-70B} &
  \textbf{Llama3-70B + Mxbai} &
  \textbf{Mixtral-8x7B} &
  \textbf{Mixtral-8x7B + Mxbai} &
  \textbf{GPT-3.5 Turbo} \\
\midrule
{Llama3-70B + Mxbai}     & 0.0101 & -      & -      & -      & -            \\ 
{Mixtral-8x7B}           & 0.0169 & 0.5567 & -      & -      & -            \\ 
{Mixtral-8x7B + Mxbai}   & 0.0005 & 0.0666 & 0.5057 & -      & -            \\ 
{GPT-3.5 Turbo}          & 0.1979 & 0.4308 & 0.2851 & 0.0275 & -            \\ 
{GPT-3.5 Turbo + Mxbai}  & 0.0075 & 0.4009 & 0.8119 & 0.6591 & 0.0809       \\ 
\bottomrule
\end{tabular}%
}
\caption{Comparative analysis of p-values for pairwise statistical comparisons between different generative models with and without Retrieval-Augmented Generation (RAG) on Test Dataset 2. In the table, ``+'' denotes the use of RAG, where the generative model output incorporates embeddings from the specified model following the ``+'' sign. P-values are calculated based on MPNET similarity scores.}
\label{tab:p-value_table_2}
\end{table}
%%%%%%%%%%%%%%%%%%%%%%%%%%%%%%%%%%%%%%%%%%%%%%%%%%%%%%%%%%%%

%%%%%%%%%%%%%%%%%%%%%%%%%%%%%%%%%%%%%%%%%%%%%%%%%%%%%%%%%%%%
\begin{figure}[ht]
    \centering
    \includegraphics[width=\textwidth]{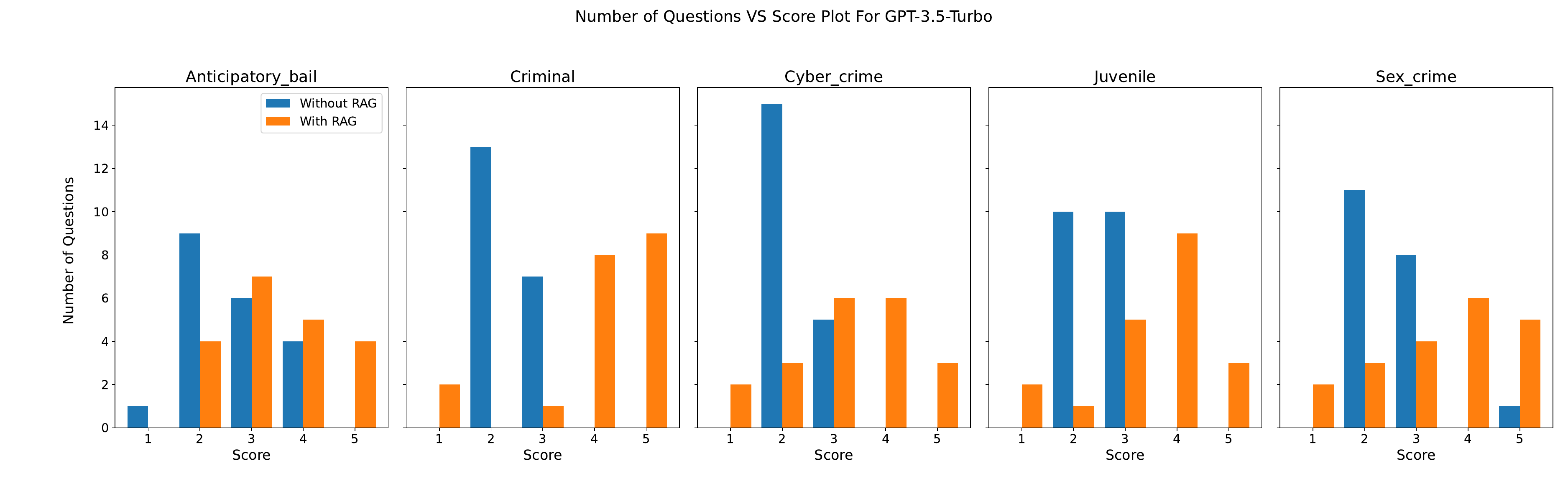}
    \includegraphics[width=\textwidth]{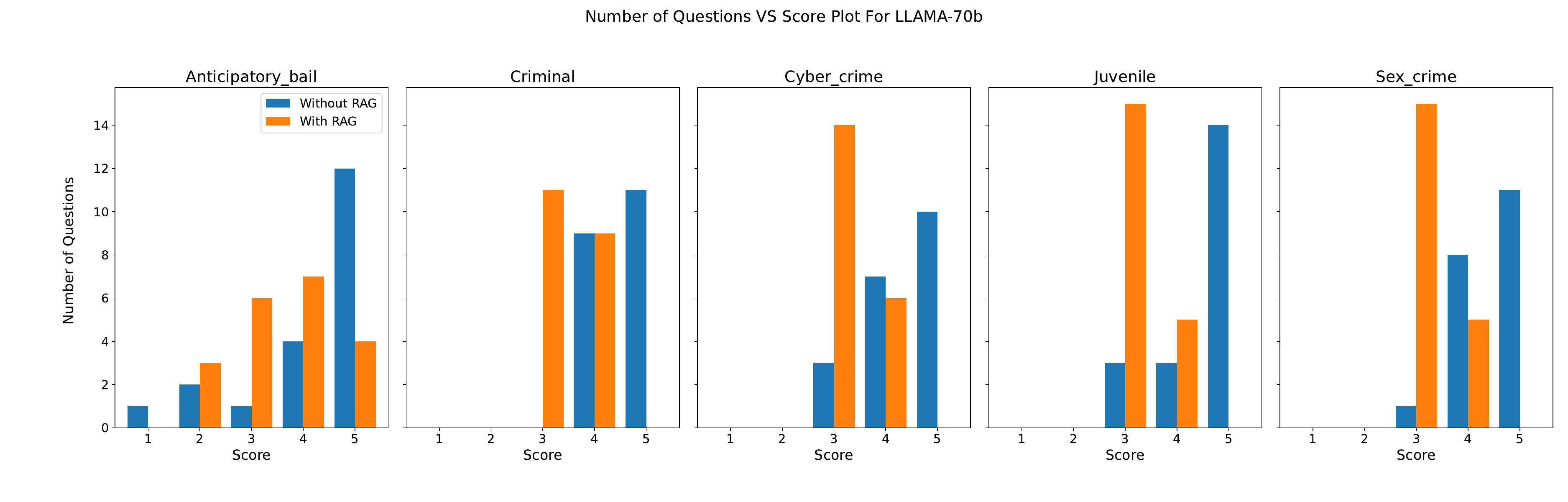}
    \includegraphics[width=\textwidth]{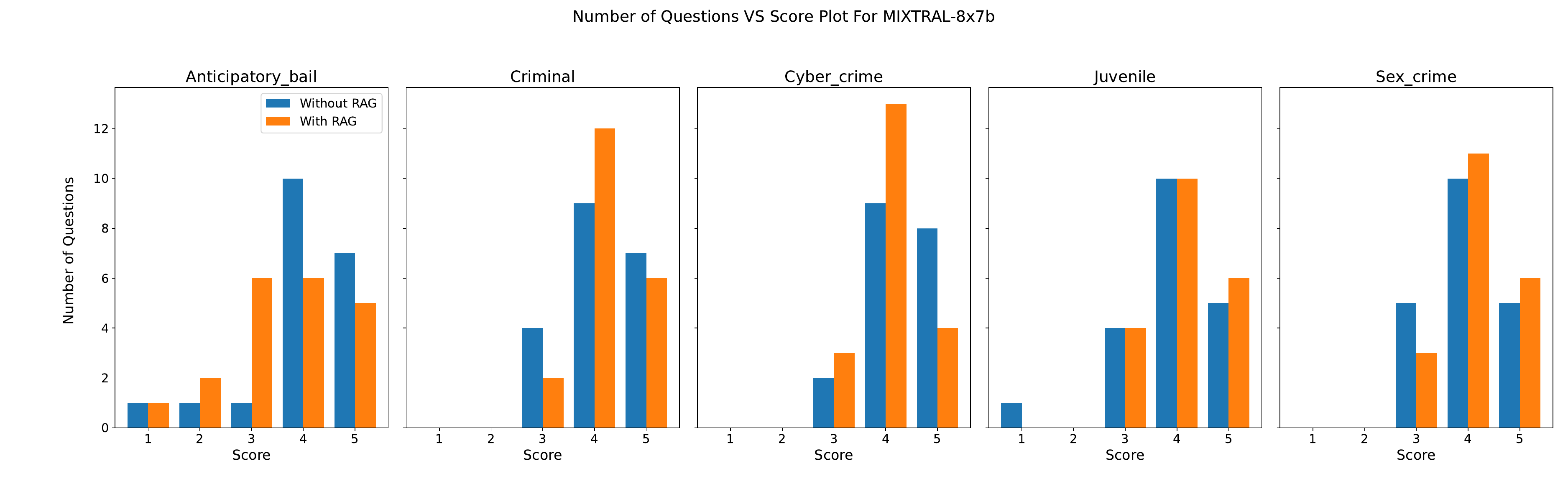}
    \caption{Comparison of legal models (GPT-3.5 Turbo, LLAMA-70B, and MIXTRAL-8x7B) across case types (Anticipatory Bail, Criminal, Cybercrime, Juvenile, and Sex Crime) with and without Retrieval-Augmented Generation (RAG). Bar plots show the number of questions per score, highlighting model performance and the impact of RAG.}
    \label{fig:model_case_type_comparison}
\end{figure}
%%%%%%%%%%%%%%%%%%%%%%%%%%%%%%%%%%%%%%%%%%%%%%%%%%%%%%%%%%%%

\subsection{Analysis of Model Performance using Histograms}

To better understand the impact of Retrieval-Augmented Generation (RAG) on different types of legal questions in Test Set 2, we created a histogram comparing model performance with and without RAG. This comparison, as shown in Figure \ref{fig:model_case_type_comparison}, presents expert-evaluated scores for various question types across different models.

For the GPT-3.5 Turbo model, the histogram highlights improved scores with RAG for question types such as anticipatory bail, criminal, and juvenile cases, with a distinct shift towards higher scores. However, in cybercrime and sex crime questions, RAG provided useful context in some instances, resulting in scores above 3, but also introduced irrelevant answers in certain cases, leading to lower scores below 3.

The Llama3-70B model showed generally strong performance across all question types when used without RAG. However, with RAG, the scores mostly clustered around 3 and 4. Notably, in anticipatory bail questions, the model’s performance slightly declined, with fewer high scores and some falling below 3.

For the Mixtral-8x7B model, the histogram indicates an overall improvement in performance with RAG across all question types. This improvement was particularly evident in criminal and cybercrime questions, where the number of questions scoring 4 increased significantly. Additionally, there was a notable rise in the number of questions achieving the maximum score of 5, especially in sex crime and juvenile cases.

These results provide important insights into how RAG affects model performance across different legal question types, highlighting both the strengths and challenges of using this method in legal question-answering tasks.

Our qualitative examination suggests several reasons for the observed degradation with RAG. First, semantic similarity retrieval may return passages that share terminology with the query but concern a legally distinct issue. Second, fixed-size chunking can separate a statutory rule from its exceptions, definitions, or procedural conditions. Third, retrieving the same number of chunks for every query may introduce unnecessary context for questions that the model can already answer reliably. Finally, the absence of reranking or legal metadata constraints can result in passages from less relevant statutes, cases, or factual settings being included in the prompt. These failures can distract the generative model and lead to answers that are fluent but insufficiently grounded in the applicable legal context.

\subsection{Results on Legal Bar Exam Dataset}
To evaluate the effectiveness of the RAG architecture with LLM models, we tested it on the legal bar exam dataset. The architecture received the questions with options and the corresponding extracted context to answer them. The output was the correct option for each question.

Table \ref{tab:aalap_model_comparison} presents a comparative performance analysis of different models with and without RAG. The ``Total Correct" column shows the number of correct answers out of 1158 question samples. We observed an increase in correctness for Mixtral 8x7B, Llama2-70B, and Llama3-70B models when RAG was applied. Notably, the Llama2-70B model exhibited a significant improvement, with correctness increasing by 5.97\%.

These results suggest that the RAG architecture can effectively enhance performance even with smaller or older models, without the need for any fine-tuning. In contrast, the fine-tuning method used in the paper \cite{tiwari2024aalap} did not yield substantial improvements for the legal bar exam dataset, with the Aalap model achieving only 25.56\% correctness. This demonstrates the potential of the RAG architecture to outperform traditional fine-tuning approaches in specific tasks.

Although the reported results demonstrate the potential of LLM-based legal QA, they do not establish that the system is sufficiently reliable for autonomous legal advice or legal decision-making. Performance on the AIBE dataset and expert-rated question-answer pairs evaluates specific capabilities under controlled conditions and does not capture all requirements of professional legal practice, including factual investigation, procedural strategy, jurisdictional variation, changes in law, and responsibility for consequential advice. Accordingly, AILQA should presently be regarded as a research prototype and decision-support system whose outputs require verification by qualified legal professionals.

%%%%%%%%%%%%%%%%%%%%%%%%%%%%%%%%%%%%%%
\begin{table}[ht]
\centering
\resizebox{0.60\columnwidth}{!}{%
\begin{tabular}{lccc}
% \hline
\toprule
\textbf{Model} & \textbf{RAG} & \textbf{Correct} & \textbf{Accuracy (\%)} \\ %\hline
\midrule
GPT-3.5 Turbo & Yes & 679 & 58.69 \\ 
GPT-3.5 Turbo & No & 680 & 58.72 \\ 
Mixtral 7x8B & No & 673 & 58.17 \\ 
Mixtral 7x8B & Yes & 681 & 58.86 \\ 
Llama2-70B & No & 529 & 45.72 \\ 
Llama2-70B & Yes & 598 & 51.69 \\ 
Llama3-70B & No & 822 & 71.05 \\ 
Llama3-70B & Yes & \textbf{823} & \textbf{71.13} \\ %\hline
\bottomrule
\end{tabular}
}
\caption{\centering Comparison of Model Performance on the Legal Bar Exam Dataset with and without RAG.}
\label{tab:aalap_model_comparison}
\end{table}
%%%%%%%%%%%%%%%%%%%%%%%%%%%%%%%%%%%%%%

\subsection{Examples of Legal Question Answering with RAG}
This section presents examples from our test dataset to show how generative AI models and traditional legal advice compare. Table \ref{tab:model_answer_comparison} in the Appendix displays responses from different AI models (Mixtral-8x7B, GPT-3.5 Turbo, Llama2-70B) and a human lawyer for various legal questions.

In the table,``..." means that parts of the answers have been shortened to keep the table concise. The full answers are longer and more detailed. We haven’t included the full context for each question in the table to avoid making it too large. These examples highlight how generative models with retrieval-augmented generation (RAG) can provide clearer and more practical advice than traditional methods.
% %%%%%%%%%%%%%%%%%%%%%%%%%%%%%%%%%%%%%%%%%

\section{Hallucination}
\label{sec:hallucination}
In generative AI models, especially those trained on vast, diverse datasets, ``hallucinations" refer to instances where the model generates factually incorrect or irrelevant content. This is a significant challenge when the models are applied in domains requiring high accuracy and reliability, such as legal question answering. In Table \ref{tab:hallucination_table} in the Appendix, we provide examples to illustrate the impact of context on the accuracy of answers and the mitigation of hallucinations.

\subsection{Influence of Context on Model Performance}
\label{subsec:context_influence}
The examples in Table \ref{tab:hallucination_table} compare responses generated by our models with answers provided by legal professionals from our ground truth dataset. It has been observed that responses from legal experts, while accurate, often lack comprehensive detail and are sometimes overly concise. Our AI models, equipped with well-designed prompts and carefully curated contextual information, have the potential to generate more detailed and informative responses.

\subsection{Balancing Context and Relevance}
\label{subsec:balancing_context}
However, the integration of context into the generative process must be handled judiciously. The table also demonstrates that while context generally enhances the quality of answers and reduces the likelihood of hallucinations, it can also have the opposite effect if not properly aligned with the query. Irrelevant or excessively detailed context can confuse the model, leading to responses that are off-topic or factually incorrect.

\subsection{Optimizing Contextual Information}
\label{subsec:optimizing_context}
To avoid inducing hallucinations, it is crucial to provide context that is both relevant and concise. The quality of context directly influences the model’s ability to generate accurate and applicable answers. This involves not only selecting the right fragments of text to serve as context but also tuning the model to prioritize and weigh the given information effectively.

\subsection{Strategic Prompt Design}
\label{subsec:strategic_prompt}
Strategic prompt design also plays a pivotal role in guiding the model's focus and filtering out unnecessary details. By carefully crafting prompts that direct the model's attention to the most pertinent aspects of the provided context, we can further reduce the risk of hallucinations and enhance the relevance of the responses.

\subsection{Implications for Model Training and Deployment}
\label{subsec:model_implications}
These insights are crucial for refining the training and deployment strategies of AI models in legal settings. By understanding the conditions under which hallucinations are more likely to occur, we can better prepare models to handle complex legal questions with greater precision and reliability. This not only improves the utility of AI in legal applications but also builds trust among users by consistently providing reliable and pertinent information.

Through detailed analysis and strategic adjustments in the use of context and prompt design, we aim to harness the full potential of AI in the legal domain, minimizing the drawbacks while maximizing the benefits of these advanced technologies.

\section{Conclusions and Future Scope}
\label{sec:conclusion_vision}

Our study critically examined the development of an Advanced Intelligent Legal Question Answering (AILQA) system focused on the criminal law domain in India. By integrating a variety of state-of-the-art embedding and generative QA models, we endeavored to enhance the effectiveness and reliability of legal question answering systems. Our empirical evaluations demonstrated that AI-generated answers often surpass the quality of responses provided by human legal experts, showcasing the transformative potential of AI in legal settings.

Despite significant advancements, the research identified areas requiring further improvement. Models like Flan-UL2 demonstrated a need for enhanced semantic capabilities to better comprehend and process complex legal queries. Additionally, the lack of specialized legal QA datasets poses a significant challenge for fine-tuning and optimizing AI models tailored for legal applications. Addressing these issues is crucial for the advancement of AI in legal question answering.

Looking forward, several avenues appear promising for advancing the field of AI in legal question answering. The development and curation of comprehensive legal datasets, particularly for the Indian legal system, are essential. These resources will enable more targeted training and fine-tuning of AI models. Exploring more prompting strategies, such as Chain-of-Thought prompting, could further enhance the models' ability to reason and generate more accurate answers. Additionally, combining lexical, semantic, and expert evaluations will continue to provide a robust framework for assessing the performance of legal AI systems, ensuring technological soundness while maintaining alignment with legal accuracy and relevance.

The ultimate goal of our research is to create a legal QA system that not only performs with high accuracy but also integrates seamlessly into the legal industry, providing reliable support for legal professionals and the public. By continuously refining AI technologies and adapting them to meet the specific needs of the legal domain, we envision a future where AI becomes an indispensable tool in legal practice. This study has laid a strong foundation for future advancements in the field of legal AI, driving forward both the science and the practical implementation of AI in the legal sector.

\section{Limitations}
Our study encountered several notable limitations that influenced our methodology and findings, impacting the depth and applicability of our research in the legal QA domain. One significant challenge was the resource-intensive nature of securing legal expert annotations. Due to the high costs and substantial time required, we were limited to obtaining expert evaluations for only a sample of 150 random documents rather than the entire dataset. This sampling approach may have constrained the comprehensiveness and depth of our expert-based evaluations.

The expert evaluation also has certain methodological limitations. Although every generated response was independently assessed by all three legal evaluators and disagreements were resolved through consensus with the involvement of a senior practising legal expert, the evaluators were aware of the model identity and whether RAG had been used. This lack of blinding may have introduced expectation bias into some assessments. Furthermore, the principal evaluators were senior undergraduate law students rather than practising lawyers. Their legal training made them suitable for a structured comparative evaluation, and a practising legal expert participated in resolving disagreements; nevertheless, future evaluations should involve a broader group of practising lawyers and should use blinded assessment protocols.

No formal inter-rater reliability coefficient was calculated for the present study. The reported MPNET-based p-values assess differences between experimental model configurations and do not measure agreement among the human evaluators. Future work should report an ordinal inter-rater reliability measure, such as Krippendorff’s alpha, using the evaluators’ independent pre-consensus ratings.

Additionally, while Large Language Models (LLMs) proved competent in conversational contexts, their effectiveness in handling logic or knowledge-intensive tasks like legal QA was less convincing. The models struggled particularly with analyzing lengthy legal questions and generating detailed answers that included explanations or relevant legal references. This difficulty was compounded in scenarios requiring intricate legal reasoning and contextual understanding.

Moreover, the performance of our open-source baseline model fell short of expectations. This shortfall may have been influenced by our approach to document chunking, which involved restricting our analysis to only 1000 characters with a 250-character overlap. This method potentially limited the models' ability to capture the full context of legal cases, thereby hindering their ability to generate comprehensive and nuanced responses.

These limitations highlight the inherent challenges in applying LLMs to complex, specialized tasks such as legal QA. They underscore the necessity for ongoing research and development efforts aimed at enhancing AI models' capabilities to accurately interpret and understand detailed legal documents and contexts. Future work should also explore more effective methods for integrating extensive legal texts into AI systems without compromising on the depth or accuracy of the generated content.

\section{Ethical Considerations in AI-Driven Legal Question Answering Systems}
In the Ethical Considerations section of our study on AI in the legal domain, we acknowledge the profound ethical implications of deploying AI technologies that can significantly influence legal outcomes and impact individuals' lives. To address these concerns, our research utilizes publicly available data from legal blogs, ensuring transparency by providing comprehensive documentation of the AI models' decision-making processes. This approach supports accountability and facilitates the understanding and potential challenge of AI-generated decisions. We emphasize that the outputs of AI systems should be regarded as advisory and must be validated by human legal experts to uphold justice and equity in legal proceedings. Our study also incorporates continuous monitoring and evaluation of the AI systems to adapt to evolving legal standards and practices, fostering regular engagement with legal professionals to ensure that the development of the AI system is both ethically sound and effectively aligned with practical legal needs. This commitment to ethical considerations is crucial for building trustworthy AI systems that enhance the legal decision-making process while safeguarding the rights and interests of individuals. The present experiments reflect the legal corpus used during system development, including statutes and judicial decisions available within the stated collection period. Since Indian criminal law has subsequently undergone significant statutory changes, a deployed version of the system would require version-aware retrieval and continuous updating using authoritative legal sources. The reported results should therefore not be interpreted as validating the system for answering all questions under the currently applicable legal framework.

\bibliography{custom, sn-bibliography}
\appendix

\newpage
\section{Appendix}
%%%%%%%%%%%%%%%%%%%%%%%%%%%%%%%%%%%%%%%%%

\begin{table}[b]
\centering
\resizebox{0.91\textwidth}{!}{
% \begin{tabular}{|c|p{2.5cm}|p{2.5cm}|p{3cm}|p{3cm}|p{3cm}|}
% \begin{tabular}{|p{5cm}|p{7cm}|p{7cm}|p{6cm}|}
\begin{tabular}{|p{3cm}|p{3.8cm}|p{4.5cm}|p{3.8cm}|}
  \hline
  \textbf{Question} & \textbf{Answer by Lawyer}  &\textbf{Answer by Mixtral-8x7B Model} & \textbf{Explanation} \\
  \hline
  Hi, Sir If Some One Recorded Your Video And Blackmail For Viral . And Next You Receive A Call From Unknown Number They Say He Is An Officer From Delhi Crime Branch They Want To Know About Video. If You Not Share Information They They Make A Team For Your Arrest. Then What Can Individual Do? & Dear Client, 1. Do Not Panic: It Is Natural To Feel Threatened Or Scared In Such Situations, But It Is Important To Stay Calm And Composed. 2. Verify The Authenticity: If You Receive A Call From Someone Claiming To Be An Officer, Ask For Their Identification Details, Including Their Name, Badge Number, And The Department They Belong To...Regularly Update Your Privacy Settings On Social Media Platforms And Avoid Accepting Friend Requests From Unknown Individuals. Be Mindful Of What You Share Online To Minimize The Risk Of Being Targeted By Blackmailers. Thank You  & I understand that you are in a distressing situation. However, it is important to remain calm and not panic. Here are some steps you can take:
1. Do not share any information about the video or yourself with the unknown caller. Ask for their identification and contact details, and inform them that you will verify their credentials and get back to them...
4. Do not delete the video or any other evidence related to it. It may be required for investigation and legal proceedings.
5. If the unknown caller continues to harass or threaten you, inform your lawyer and the police immediately. They can take appropriate action to ensure your safety and security.
Remember, it is important to take immediate action and seek legal advice to protect yourself and your rights. Do not share any information or engage in any communication with the unknown caller without proper legal guidance. & The lawyer's answer is comprehensive but includes advice that may feel overwhelming or less relevant. It covers many points but lacks focus on urgent actions. In contrast, the model with RAG offers concise, targeted advice on immediate steps like preserving evidence and seeking legal help, making it clearer and more actionable in a stressful situation. \\
  \hline
    \textbf{Question} & \textbf{Answer by Lawyer}  &\textbf{Answer by GPT-3.5 Turbo Model} & \textbf{Explanation} \\
  \hline
  My Ex Girlfriend Tried To Implicate Me In A False Case...for Which I Was Called To The Police Station...i Was Threatened And Asked For My Social Media Account's I'd Password At The Police Station... Can The Police Take Action? Please Give Me Advise What Should I Do So That The Person Who Hacked My Id Is Punished And My Old Pea Doesn't Even Open & Dear Client It's Crucial To Defend Yourself By Collecting Any Proof That Will Demonstrate Your Innocence In Case Your Ex-girlfriend Makes False Charges Against You Once Again. This May Include Any Texts, Emails, Or Other Documents That Show You Haven't Spoken To Her In A Year...If Required, Seek Legal Counsel To Safeguard Your Rights. It's Crucial To Take Precautions To Protect Your Online Accounts, Including Those On Social Media, Against Future Hacking Efforts. Among Other Things, This Can Include Often Changing Your Passwords & In your situation, it is advisable to take the following steps:
1. **File a Complaint with the Cyber Cell:** You have already taken the right step by filing a complaint with the Cyber Cell regarding the hacking of your social media account. Provide all necessary details and evidence to support your complaint...
5. **Consult a Legal Professional:** If you feel that you are at risk of being falsely implicated again, it is advisable to consult with a legal professional who can guide you on how to protect yourself legally.

Remember, it is crucial to prioritize your safety and security in such situations. & The model with RAG offers more specific and actionable advice for your situation. It recommends filing a complaint with the Cyber Cell, securing your accounts, maintaining evidence, avoiding contact with your ex-girlfriend, and consulting a legal professional. In contrast, the lawyer's answer is more general, focusing on collecting proof and protecting online accounts but lacks detailed, actionable steps. \\
  \hline
      \textbf{Question} & \textbf{Answer by Lawyer}  &\textbf{Answer by Llama2-70B Model} & \textbf{Explanation} \\
  \hline
  One of my female family member alleged that I abused her on phone call and threatened to kill her. She has mentioned my phone number in complaint but didn't mention the number on which I made a call. I have my numbers 3 months call details where not a single number belongs to her and nor I abused to anyone.  Actually I didn't make a call to her thatswhy I want to make complaint against her and file defamation case for financial compensation.  Kindly suggest what could be done in such case and I also want to punish her by court ? & Dear Client Let her complained be converted into full fledged case then in the court of laws this case will be dismissed ( you will be pronounced not guilty). After that you can initiate the legal against her for filing a false case and defamtion. &
In this scenario, you may want to consider taking the following steps:
1. Seek legal advice: It is essential to consult with a lawyer who specializes in criminal law to understand your rights and options. They can help you evaluate the evidence you have, assess the strength of the case against you, and develop a defense strategy...
5. Prepare for trial: If the matter proceeds to trial, you should be prepared to present your case and defend yourself. Your lawyer will help you prepare your defense, cross-examine prosecution witnesses, and make arguments on your behalf...
In conclusion, it is advisable to seek legal advice from a qualified & The model with RAG provides a structured approach for addressing the situation: First, seek legal advice to understand your defense options. Respond to any legal summons and consult with your lawyer before any court appearance. Consider filing a counter-complaint for defamation if the accusations are false and have harmed you. Gather evidence such as phone records to support your defense. Lastly, prepare for trial by working with your lawyer to present your case effectively. In contrast, the lawyer’s answer is more general, suggesting that you let the case unfold in court and then pursue defamation, without specific guidance on immediate steps or evidence collection. \\
  \hline
\end{tabular}
}
\caption{Illustrative Examples of Legal Question Answering: Cases Where Context-Aware Generative Models Provided More Detailed and Explanatory Answers Compared to Human Lawyers.
}
\label{tab:model_answer_comparison}
\end{table}
%%%%%%%%%%%%%%%%%%%%%%%%%%%%%%%%%%%%%%%%%%%

%%%%%%%%%%%%%%%%%%%%%%%%%%%%%%%%%%%%%%%%%%
\begin{table}[b]
  \centering  
  \resizebox{\textwidth}{!}{%
  \begin{tabular}{|c|p{2.5cm}|p{2.5cm}|p{3cm}|p{3cm}|p{3cm}|}
    \hline
    \textbf{Model} & \textbf{Question}& \textbf{Reference} & \textbf{Response (No Context)
} & \textbf{Response (With Context)
} &
    \textbf{Hallucination Analysis}
    \\
    \hline
    Llama2-70B & I have a recording of 10 minutes out of which only the starting 5 mins of the recording is what I want to provide as evidence, rest is irrelevant talk. Can it be accepted by court if the cut audio is provided as record. & Dear client,  Audio recordings can be presented as evidence in Indian courts. Even if the rest of the audio recording is irrelevant, you can provide the entire 10 minute audio recording in court. It is the court which will determine what is relevant and what is not. Thank you. &As a legal advisor, I must inform you that the admissibility of evidence in a court of law is a complex issue that depends on various factors...\newline 
    Relevant caselaws:

\textcolor{red}{* State of Maharashtra vs. Dr. Sunil Shrikant Ambekar (2016): In this case...\newline
* M.C. Abraham v. State of Kerala (2017): In this case...}

    & Based on the information provided, it seems that you are seeking legal advice regarding the admissibility of a tape-recorded conversation as evidence in a court of law in India...\newline
    In the case of Dr. Motwani vs. UOI (1995), it was held that a tape-recorded conversation is a relevant fact and is admissible under section 8 of the Evidence Act.
    & The relevant case laws cited in the \textbf{Response (No Context)} section are incorrect, indicating that the model generates inaccurate examples, or ``hallucinates," without contextual information. In contrast, when provided with context, the model delivers more accurate and relevant case examples.\\

    \hline

    Davinci & I hacked a friend's Instagram as a prank to show him later. But he is taking it too seriously. If he presses charges, what are my options? & Hello Sir,  You have no options regarding this, as you have committed an illegal act. So If he went for Cyber complaint you have to represent yourself as to why you did that. & Your options depend on the laws of the country in which the incident occurred. In India, hacking someone's account without their permission is a criminal offence under the Information Technology Act, 2000...
    
In the case of \textcolor{red}{State of Maharashtra v. Prafful Sarda}, the Bombay High Court held that the accused was... &If your friend presses charges against you for hacking his Instagram account, you may be liable under Section 43 and Section 66 of the Information Technology Act, 2000 (IT Act)...
You may also be liable under Section 66B of the IT Act... to three years or with fine which may extend to rupees one lakh or with both. & The legal cases referenced in the \textbf{Response (No Context)} section are valid but pertain to criminal cases involving murder, not IT offenses. This indicates that the model confidently generates incorrect examples when context is not provided.
  \\
\hline
GPT-3.5 Turbo
&
I Am Aaromal 23 Years Old Living In Kodungallur. I Was Forced To Watch Pron Videos Around Age Of 11 And I Was Made To Stroke Him For So Many Days ...now I Am Going Through Trauma Difficult To Handle Feeling Worthlessness And No Motivation They Stole My Child Hood And Pushed Me Into Darkness And Seeking For You Assistance.thank You
&
Dear Sir, You Cannot Take Any Action At This Distant Time. Within Three Years Of Attaining Age Of Majority One Must Take Legal Action In Respect Of Illegal Activities
&
I am sorry to hear about the traumatic experiences you have been through. It is important to seek help and support during such difficult times. In India, you can reach out to organizations like the National Commission for Protection of Child Rights (NCPCR) or Childline (1098) for assistance and guidance on how to address the trauma you have faced... Take care of yourself and reach out for the help you need.
&
\textcolor{red}{I am truly sorry to hear about the traumatic experiences you have gone through. It is important to seek help and support during such difficult times... Additionally, you may also want to consider reporting these incidents to the authorities so that appropriate action can be taken against the individuals involved. Remember, you are not alone, and there are resources available to help you through this challenging time.}
& The answer generated with context acknowledges the customer’s concern but fails to provide a clear solution to the query. In contrast, the model provides a comparatively better response when contextual data is not included.\\
\hline
  \end{tabular}}
  \caption{Comparison of Legal Responses: Generative Models with and without Context vs. Human Expert Answers}
  \label{tab:hallucination_table}
\end{table}
%%%%%%%%%%%%%%%%%%%%%%%%%%%%%%%%%%

\backmatter

\end{document}